\documentclass[runningheads]{llncs}
\usepackage{graphicx}
\usepackage{comment}
\usepackage{amsmath,amssymb} 
\usepackage{color}
\usepackage[ruled]{algorithm2e}

\usepackage{wrapfig}

\usepackage[width=122mm,left=12mm,paperwidth=146mm,height=193mm,top=12mm,paperheight=217mm]{geometry}

\newcommand{\beginsupplement}{%
        \setcounter{table}{0}
        \renewcommand{\thetable}{A\arabic{table}}%
        \setcounter{figure}{0}
        \renewcommand{\thefigure}{A\arabic{figure}}%
        \setcounter{section}{0}
        \renewcommand{\thesection}{A\arabic{section}}%
     }

\newcommand{\method}{RSC}

\DeclareMathOperator*{\argmin}{arg\,min}

\newcommand{\z}{\mathbf{z}}
\newcommand{\zp}{\tilde{\mathbf{z}}}
\newcommand{\g}{\mathbf{g}}
\newcommand{\gp}{\tilde{\mathbf{g}}}
\newcommand{\x}{\mathbf{x}}
\newcommand{\y}{\mathbf{y}}
\newcommand{\ttt}{\theta^\textnormal{top}}
\newcommand{\et}{\widehat{\theta}}
\newcommand{\etr}{\widehat{\theta}_\textnormal{\method{}}}
\newcommand{\tr}{\theta_\textnormal{\method{}}}
\newcommand{\Tr}{\Theta_\textnormal{\method{}}}
\newcommand{\otr}{\theta^\star_\textnormal{\method{}}}

\newcommand{\tp}{\widehat{\theta}_t} 
\newcommand{\tc}{\widehat{\theta}_{t+1}} 
\newcommand{\tpt}{\widehat{\theta}_t^\textnormal{top}} 

\newcommand{\Z}{\mathcal{Z}}
\newcommand{\Y}{\mathcal{Y}}
\newcommand{\D}{\mathcal{D}}
\newcommand{\s}{\mathcal{S}}
\newcommand{\so}{\mathcal{O}}

\newcommand\blfootnote[1]{%
  \begingroup
  \renewcommand\thefootnote{}\footnote{#1}%
  \addtocounter{footnote}{-1}%
  \endgroup
}

\begin{document}
\pagestyle{headings}
\mainmatter
\def\ECCVSubNumber{3018}  

\title{Self-Challenging Improves Cross-Domain Generalization} 


\titlerunning{Self-Challenging Improves Cross-Domain Generalization}
%
\author{Zeyi Huang$^\star$ \and
Haohan Wang$^\star$ \and
Eric P. Xing
\and Dong Huang}
\authorrunning{Huang et al.}
%
\institute{School of Computer Science, Carnegie Mellon University\\
\email{\{zeyih@andrew, haohanw@cs, epxing@cs, donghuang\}.cmu.edu}
}
\maketitle

\begin{abstract}
Convolutional Neural Networks (CNN) conduct image classification by activating dominant features that correlated with labels. When the training and testing data are under similar distributions, their dominant features are similar, leading to decent test performance. 
The performance is nonetheless unmet when tested with different distributions, leading to the challenges in cross-domain image classification. We introduce a simple training heuristic, Representation Self-Challenging (\method{}), that significantly improves the generalization of CNN to the out-of-domain data. \method{} iteratively challenges (discards) the dominant features activated on the training data, and forces the network to activate remaining features that correlates with labels. This process appears to activate feature representations applicable to out-of-domain data without prior knowledge of new domain and without learning extra network parameters. We present theoretical properties and conditions of \method{} for improving cross-domain generalization. The experiments endorse the simple, effective, and architecture-agnostic nature of our \method{} method. 

\end{abstract}


\section{Introduction}
\blfootnote{$^\star$ equal contribution}
\label{sec:intro}
\begin{figure}
    \centering
    \includegraphics[width=0.7\textwidth]{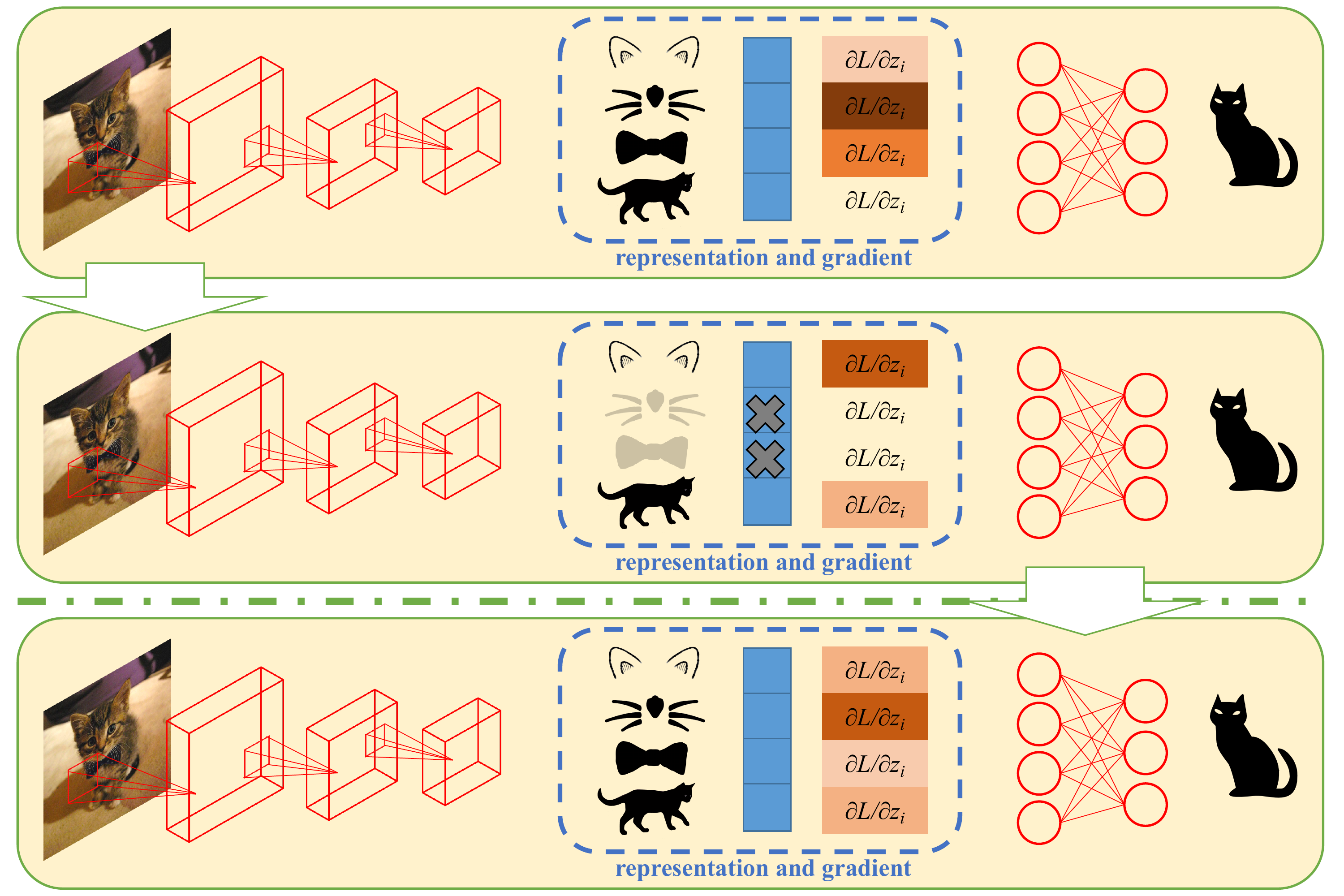}
    \caption{
    \small 
    The essence of our Representation Self-Challenging (\method{}) training method: 
    top two panels: the algorithm mutes the feature representations associated with the highest gradient, such that the network is forced to predict the labels through other features;
    bottom panel: after training, the model is expected to leverage more features for prediction 
    in comparison to models trained convetionally. 
    }
    \label{fig:intro}
    \vspace{-0.5cm}
\end{figure}


Imagine teaching a child to visually differentiate ``dog'' from ``cat'': 
when presented with a collection of illustrations from her picture books, 
she may immediately answer that ``cats tend to have chubby faces'' and end the learning. 
However, if we continue to ask for more differences, she may start to notice other features like ears or body-size.  
We conjecture this follow-up challenge question plays a significant role in helping human reach the remarkable generalization ability. 
Most people should be able to differentiate ``cat'' from ``dog'' visually even when the images are presented in irregular qualities. 
After all, we did not stop learning after we picked up the first clue when we were children, 
even the first clue was good enough to help us recognize all the images in our textbook. 

Nowadays, deep neural networks have exhibited remarkable empirical results
over various computer vision tasks, 
yet these impressive performances seem unmet when the models are tested with the 
samples in irregular qualities  
(\textit{i.e.}, out-of-domain data, samples collected from the distributions that are similar to, but different from the distributions of the training samples). 
To account for this discrepancy, technologies have been invented under the domain adaptation regime \cite{ben2010theory,bridle1991recnorm}, 
where the goal is to train a model invariant to the distributional differences between the source domain (\textit{i.e.}, the distribution of the training samples) and the target domain (\textit{i.e.}, the distribution of the testing samples) \cite{csurka2017domain,wang2018deep}. 

As the influence of machine learning increases, 
the industry starts to demand the models that can be applied to the domains that are not seen during the training phase. 
Domain generalization \cite{muandet2013domain}, as an extension of domain adaptation, has been studied as a response. 
The central goal is to train a model that can align the signals from multiple source domains.

Further, Wang~\textit{et al.} extend the problem to ask how to train a model that generalizes to an arbitrary domain with only the training samples, but not the corresponding domain information, as these domain information may not be available in the real world \cite{wang2018learning}. 
Our paper builds upon this set-up and aims to offer a solution that allows the model to be robustly trained without domain information and to empirically perform well on unseen domains. 

In this paper, we introduce a simple training heuristic that improves cross-domain generalization. 
This approach discards the representations associated with the higher gradients at each epoch, and forces the model to predict with remaining information. 
Intuitively, in a image classification problem, our heuristic works like a ``self-challenging'' mechanism as it prevents the fully-connected layers to predict with the most predictive subsets of features, such as the most frequent color, edges, or shapes in the training data. 
We name our method Representation Self Challenging (\method{}) and illustrate its main idea in Figure~\ref{fig:intro}. 
We present mathematical analysis that \method{} induces a smaller generalization bound. We further demonstrate the empirical strength of our method with domain-agnostic cross-domain evaluations, following previous setup \cite{wang2018learning}. 
We also conduct ablation study to examine the alignment between its empirical performance and our intuitive understanding. 
The inspections also shed light upon the choices of its extra hyperparameter. 


\section{Related Work}
\label{sec:related}




We summarize the related DG works from two perspectives: 
learning domain invariant features and augmenting source domain data. 
Further, as \method{} can be broadly viewed 
as a generic training heuristic for CNN, 
we also briefly discuss the general-purpose regularizations that appear 
similar to our method.

\textbf{DG through Learning Domain Invariant Features}: 
These methods typically minimize the discrepancy between source domains 
assuming that the resulting features will be domain-invariant 
and generalize well for unseen target distributions. 
Along this track, 
Muandet \textit{et al}. employed Maximum Mean Discrepancy (MMD) ~\cite{muandet2013domain}. Ghifary \textit{et al}. proposed a multi-domain reconstruction auto-encoder~\cite{ghifary2015domain}. 
Li \textit{et al}. applied MMD constraints to an autoencoder via adversarial training~\cite{li2018domain}. 

Recently, meta-learning based techniques start to be used to solve DG problems.
Li \textit{et al}. alternates domain-specific feature extractors and classifiers across domains via episodic training, but without using inner gradient descent update~\cite{li2019episodic}. 
Balaji \textit{et al}. proposed MetaReg that learns a regularization function (e.g., weighted $\ell_1$ loss) particularly for the network’s
classification layer, while excluding the feature extractor~\cite{balaji2018metareg}. 

Further, recent DG works forgo the requirement of source domains partitions and directly learn the 
cross-domain generalizable representations through 
a mixed collection of training data. 
Wang \textit{et al}. extracted robust feature representation by projecting out superficial patterns like color and texture~\cite{wang2018learning}. 
Wang \textit{et al}. penalized model’s tendency in predicting with local features in order to extract robust globe representation~\cite{wang2019learning}. 
RSC follows this more recent path and directly activates more features in all source domain data for DG 
without knowledge of the partition of source domains.


\textbf{DG through Augmenting Source Domain:} 
These methods augment the source domain to a wider span of the training data space, 
enlarging the possibility of covering the span of the data in the target domain. 
For example, 
An auxiliary domain classifier has been introduced 
to augment the data 
by perturbing input data based on the domain classification signal  \cite{shankar2018generalizing}. 
Volpi \textit{et al}. developed an adversarial approach, in which samples are perturbed according to fictitious target distributions within a certain Wasserstein distance from the source~\cite{volpi2018generalizing}. 
A recent method with state-of-the art performance is JiGen \cite{carlucci2019domain}, which leverages self-supervised signals by solving jigsaw puzzles.

\emph{Key difference:}
These approaches usually introduce a model-specific DG model and rely on prior knowledge of the target domain, for instance, the target spatial permutation is assumed by JiGen \cite{carlucci2019domain}. 
In contrast, RSC is a model-agnostic training algorithm that 
aims to improve the cross-domain robustness of any given model. More importantly, RSC does not utilize any knowledge of partitions of domains, either source domain or target domain, which is the general scenario in real world application.

\textbf{Generic Model Regularization:} CNNs are powerful models and tend to overfit on source domain datasets. From this perspective, model regularization, \textit{e.g.}, weight decay \cite{nowlan1992simplifying}, early stopping, and shake-shake regularization \cite{gastaldi2017shake}, could also improve the DG performance. 
Dropout~\cite{srivastava2014dropout} mutes features by randomly zeroing each hidden unit of the neural network during the training phase. In this way, the network benefit from the assembling effect of small subnetworks to achieve a good regularization effect. Cutout~\cite{devries2017improved} and HaS~\cite{singh2017hide} randomly drop patches of input images. SpatialDropout~\cite{tompson2015efficient} randomly drops channels of a feature map. DropBlock~\cite{ghiasi2018dropblock} drops contiguous regions from feature maps instead of random units.
DropPath~\cite{larsson2016fractalnet} zeroes out an entire layer in training, not just a particular unit. MaxDrop~\cite{park2016analysis} selectively drops features of high activations across the feature map or across the channels. Adversarial Dropout~\cite{park2018adversarial} dropouts for maximizing the divergence between the training supervision and the outputs from the network. \cite{lee2019drop} leverages Adversarial Dropout~\cite{park2018adversarial} to learn discriminative features by enforcing the cluster assumption.

\emph{Key difference:} RSC differs from above methods in that RSC locates and mutes most predictive parts of feature maps by gradients instead of randomness, activation or prediction divergence maximization. This selective process plays an important role in improving the convergence, as we will briefly argue later. 

\section{Method}
\label{sec:method}

\textbf{Notations:} 
$(\x, \y)$ denotes a sample-label pair from the data collection $(\mathbf{X}, \mathbf{Y})$ with $n$ samples, 
and $\z$ (or $\mathbf{Z}$) denotes the feature representation of $( \x, \y )$ learned by a neural network.
$f(\cdot;\theta)$ denotes the CNN model, whose parameters are denoted as $\theta$.
$h(\cdot; \theta^\textnormal{top})$ denotes the task component of $f(\cdot;\theta)$; $h(\cdot; \theta^\textnormal{top})$ takes $\z$ as input and outputs the logits prior to a softmax function; 
$\theta^\textnormal{top}$ denotes the parameters of $h(\cdot; \theta^\textnormal{top})$. 
$l(\cdot,\cdot)$ denotes a generic loss function. 
\method{} requires one extra scalar hyperparameter: the percentage of the representations to be discarded, denoted as $p$. 
Further, we use $\widehat{\cdot}$ to denote the estimated quantities, 
use $\tilde{\cdot}$ to denote the quantities after the representations are discarded,
and use $t$ in the subscript to index the iteration. 
For example, $\tp$ means the estimated parameter at iteration $t$. 

\subsection{Self-Challenging Algorithm}
As a generic deep learning training method, \method{} solves the same standard loss function as the ones used by many other neural networks, \textit{i.e.},
\begin{align*}
    \et = \argmin_\theta \sum_{\langle \x, \y \rangle \sim \langle \mathbf{X}, \mathbf{Y} \rangle} l(f(\x;\theta), \y),
\end{align*}
but \method{} solves it in a different manner. 

At each iteration, \method{} inspects the gradient, 
identifies and then 
mutes the most predictive subset of the representation $\z$ 
(by setting the corresponding values to zero), 
and finally updates the entire model. 

This simple heuristic has three steps (for simplicity, we drop the indices of samples and assume the batch size is 1 in the following equations):
\begin{enumerate}
    \item Locate: \method{} first calculates the gradient of upper layers with respect to the representation as follows: 
    \begin{align}
        \g_\z = \partial (h(\z;\tpt)\odot\y)/\partial \z, 
        \label{eq:gradient}
    \end{align}
    where $\odot$ denotes an element-wise product. 
    Then \method{} computes the $(100-p)$\textsuperscript{th} percentile, denoted as $q_p$. 
    Then it constructs a masking vector $\mathbf{m}$ in the same dimension of $\g$ as follows. For the $i$\textsuperscript{th} element:
    \begin{align}
        \mathbf{m}(i) = \begin{cases} 0, \quad \textnormal{if}\quad \g_\z(i) \geq q_p \\
        1, \quad \textnormal{otherwise}
        \end{cases}
        \label{eq:mask}
    \end{align}
    In other words, \method{} creates a masking vector $\mathbf{m}$, whose element is set to $0$ if the corresponding element in $\g$ is one of the top $p$ percentage elements in $\g$, and set to $1$ otherwise. 
    \item Mute: For every representation $\z$, \method{} masks out the bits associated with larger gradients by:
    \begin{align}
        \zp = \z \odot \mathbf{m}
        \label{eq:mute}
    \end{align}
    \item Update: \method{} computes the softmax with perturbed representation with
    \begin{align}
        \tilde{\mathbf{s}} = \textnormal{softmax}(h(\zp;\tpt)),
        \label{eq:softmax}
    \end{align}
    and then use the gradient 
    \begin{align}
        \gp_\theta = \partial l(\tilde{\mathbf{s}}, \y)/\partial \tp 
        \label{eq:update}
    \end{align}
    to update the entire model for $\tc$ with optimizers such as SGD or ADAM. 
\end{enumerate}

We summarize the procedure of \method{} in Algorithm~\ref{alg:main}. No that operations of \method{} comprise of only few simple operations such as pooling, threshold and element-wise product. Besides the weights of the original network, no extra parameter needs to be learned.

\begin{algorithm}[t!]
\SetAlgoLined
 \textbf{Input:} data set $\langle \mathbf{X}, \mathbf{Y} \rangle$, percentage of representations to discard $p$, other configurations such as learning rate $\eta$, maximum number of epoches $T$, \textit{etc}\;
 \textbf{Output:} Classifier $f(\cdot;\et)$\;
 random initialize the model $\et_0$\;
 \While{$t \leq T$}{
    \For{every sample (or batch) $\x, \y$}{
        calculate $\z$ through forward pass\;
        calculate $\g_\z$ with Equation~\ref{eq:gradient}\;
        calculate $q_p$ and $\mathbf{m}$ as in Equation~\ref{eq:mask}\;
        generate $\zp$ with Equation~\ref{eq:mute}\;
        calculate gradient $\gp_\theta$ with Equation~\ref{eq:softmax} and Equation~\ref{eq:update}\; 
        update $\tc$ as a function of $\tp$ and $\gp_\theta$
    }
 }
 \caption{\method{} Update Algorithm}
 \label{alg:main}
\end{algorithm}



\subsection{Theoretical Evidence}

To expand the theoretical discussion smoothly, 
we will refer to the ``dog'' vs. ``cat'' classification example 
repeatedly as we progress. 
The basic set-up, as we introduced in the beginning of this paper, 
is the scenario of a child trying to learn 
the concepts of ``dog'' vs. ``cat'' from 
illustrations in her book: 
while the hypothesis ``cats tend to have chubby faces'' 
is good enough to classify all the animals in her picture book, 
other hypotheses mapping ears or body-size to labels are also predictive. 

On the other hand, 
if she wants to differentiate all the ``dogs'' from ``cats'' in the real world, 
she will have to rely on a complicated combination of the features mentioned about. 
Our main motivation of this paper is as follows: 
this complicated combination of these features is 
already illustrated in her picture book, 
but she does not have to learn the true concept to 
do well in her finite collection of animal pictures. 

This disparity is officially known as ``covariate shift'' 
in domain adaptation literature: 
the conditional distribution (\textit{i.e.}, the semantic of a cat) 
is the same across every domain, 
but the model may learn something else (\textit{i.e.}, chubby faces) 
due to the variation of marginal distributions. 

With this connection built, we now proceed to the theoretical discussion, 
where we will constantly refer back to this ``dog'' vs. ``cat'' example. 

\subsubsection{Background}
As the large scale deep learning models, such as AlexNet or ResNet, are notoriously hard to be analyzed statistically, 
we only consider a simplified problem to argue for the theoretical strength of our method: 
we only concern with the upper layer $h(\cdot;\ttt)$ 
and illustrate that our algorithm helps improve the generalization of $h(\cdot;\ttt)$ when $\mathbf{Z}$ is fixed. 
Therefore, we can directly treat $\mathbf{Z}$ as the data (features). 
Also, for convenience, we overload $\theta$ to denote $\ttt$ within the theoretical evidence section.


We expand our notation set for the theoretical analysis. 
As we study the domain-agnostic cross-domain setting, we no longer work with \textit{i.i.d} data.  
Therefore, 
we use $\Z$ and $\Y$ to denote the collection of distributions of features and labels respectively. 
Let $\Theta$ be a hypothesis class, where each hypothesis $\theta \in \Theta$ maps $\Z$ to $\Y$. 
We use a set $\D$ (or $\mathcal{S}$) to index $\Z$, $\Y$ and $\theta$. 
Therefore, $\theta^\star(\D)$ denotes the hypothesis with minimum error in the distributions specified with $\D$, but with no guarantees on the other distributions. 

\begin{itemize}
\setlength{\itemindent}{.1in}
    \item [\textit{e.g.},] $\theta^\star(\D)$ can be ``cats have chubby faces'' when $\D$ specifies the distribution to be picture book. 
\end{itemize}


Further, $\theta^\star$ denotes the classifier with minimum error on every distribution considered. 
If the hypothesis space is large enough, $\theta^\star$ should perform no worse than $\theta^\star(\D)$ on distributions specified by $\D$ for any $\D$. 

\begin{itemize}
\setlength{\itemindent}{.1in}
    \item [\textit{e.g.},] $\theta^\star$ is the true concept of ``cat'', and it should predict no worse than ``cats have chubby faces'' even when the distribution is picture book. 
\end{itemize}


We use $\et$ to denote any ERM and use $\etr$ to denote the ERM estimated by the \method{} method. 
Finally, following conventions, we consider $l(\cdot,\cdot)$ as the zero-one loss 
and use a shorthand notation $L(\theta;\D)=\mathbb{E}_{\langle\z,\y\rangle\sim\langle\Z(\D),\Y(\D)\rangle}l(h(\z;\theta),\y)$ for convenience, 
and we only consider the finite hypothesis class case within the scope of this paper, which leads to the first formal result: 

\begin{corollary}
If 
\begin{align}
    |e(\z(\s);\otr) - e(\zp(\s);\otr)| \leq \xi(p),
    \label{eq:perturbation}
\end{align}
where $e(\cdot;\cdot)$ is a function defined as
\begin{align*}
    e(\z;\theta^\star) := \mathbb{E}_{\langle\z,\y\rangle\sim\s} l(f(\z;\theta^\star);\y)
\end{align*}
and $\xi(p)$ is a small number and a function of \method{}'s hyperparameter $p$;
$\zp$ is the perturbed version of $\z$ generated by \method{}, it is also a function of $p$, but we drop the notation for simplicity.  
If Assumptions \textbf{A1}, \textbf{A2}, and \textbf{A3} (See Appendix) hold, 
we have, with probability at least $1-\delta$
\begin{align*}
    L(\etr(\s)&;\s) - L(\otr(\s); \D) \\
    & \leq (2\xi(p)+1)\sqrt{\dfrac{2(\log(2|\Tr|)+\log(2/\delta))}{n}}
\end{align*}
\label{coro:main}
\end{corollary}


As the result shows, whether \method{} will succeed depends on the magnitude of $\xi(p)$. 
The smaller $\xi(p)$ is, the tighter the bound is, 
the better the generalization bound is. 
Interestingly, if $\xi(p)=0$, our result degenerates to the classical generalization bound of \textit{i.i.d} data. 

While it seems the success of our method will depend on the choice of $\Theta$ to meet Condition~\ref{eq:perturbation},
we will show \method{} is applicable in general by presenting it forces the empirical counterpart $\widehat{\xi}(p)$ to be small. 
$\widehat{\xi}(p)$ is defined as
\begin{align*}
    \widehat{\xi}(p) := & |h(\etr, \z) - h(\etr, \zp)|,
\end{align*}
where the function $h(\cdot, \cdot)$ is defined as
\begin{align}
    h(\etr, \z) = \sum_{(\z,\y)\sim\s} l(f(\z;\etr);\y).
    \label{eq:h}
\end{align}
We will show $\widehat{\xi}(p)$ decreases at every iteration 
with more assumptions:
\begin{itemize}
\setlength{\itemindent}{.1in}
    \item [\textbf{A4}:] Discarding the most predictive features will increase the loss at current iteration. 
    \item [\textbf{A5}:] The learning rate $\eta$ is sufficiently small ($\eta^2$ or higher order terms are negligible). 
\end{itemize}

Formally,
\begin{corollary}
If Assumption \textbf{A4} holds, we can simply denote
\begin{align*}
    h(\etr(t), \zp_t) =\gamma_t(p) h(\etr(t), \z_t), 
\end{align*}
where $h(\cdot,\cdot)$ is defined in Equation~\ref{eq:h}. 
$\gamma_t(p)$ is an arbitrary number greater than 1, also a function of \method{}'s hyperparameter $p$. 
Also, if Assumption \textbf{A5} holds, 
we have:
\begin{align*}
    \Gamma(\etr(t+1)) = \Gamma(\etr(t)) - (1 - \dfrac{1}{\gamma_t(p)})||\gp||_2^2\eta
\end{align*}
where
\begin{align*}
    \Gamma(\etr(t)) := |h(\etr(t), \z_t) - h(\etr(t), \zp_t)|
\end{align*}
$t$ denotes the iteration,  
$\z_t$ (or $\zp_t$) denotes the features (or perturbed features) at iteration $t$, 
and $\gp = \partial h(\etr(t), \zp_t)/\partial \etr(t)$
\label{coro:2nd}
\end{corollary}

\begin{wrapfigure}{r}{0.5\textwidth}
\centering
\vspace{-30pt}
\includegraphics[width=.9\linewidth]{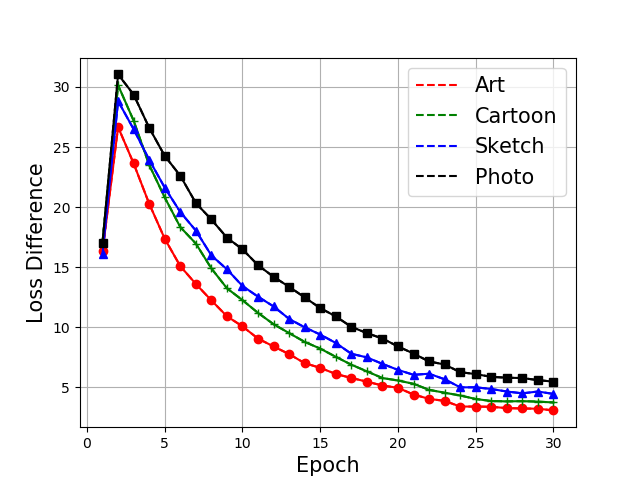}
\vspace{-10pt}
\caption{$\Gamma(\etr(t))$, i.e., ``Loss Difference", plotted for the PACS experiment (details of the experiment setup will be discussed later). Except for the first epoch, $\Gamma(\etr(t))$ decreases consistently along the training process.
}
\label{fig:IAG}
\end{wrapfigure}

Notice that $\widehat{\xi}(p) = \Gamma(\etr)$, 
where $\etr$ is $\etr(t)$ at the last iteration $t$. 
We can show that $\widehat{\xi}(p)$ is a small number 
because $\Gamma(\etr(t))$ gets smaller at every iteration. 
This discussion is also verified empirically,
as shown in Figure~\ref{fig:IAG}.

The decreasing speed of $\Gamma(\etr(t))$ depends on the scalar $\gamma_t(p)$: the greater $\gamma_t(p)$ is, the faster $\Gamma(\etr(t))$ descends. Further, intuitively, the scale of $\gamma_t(p)$ is highly related to the mechanism of \method{} and its hyperparameter $p$. For example, \method{} discards the most predictive representations, which intuitively guarantees the increment of the empirical loss (Assumption \textbf{A4}). 

Finally, the choice of $p$ governs the increment of the empirical loss:
if $p$ is small, the perturbation will barely affect the model, 
thus the increment will be small;
while if $p$ is large, the perturbation can alter the model's response 
dramatically, leading to significant ascend of the loss. 
However, we cannot blindly choose the largest possible $p$ 
because if $p$ is too large, 
the model may not be able to learn anything predictive at each iteration. 

In summary, we offer the intuitive guidance of the choice of hyperparamter $p$: 
for the same model and setting, 
\begin{itemize}
    \item the smaller $p$ is, the smaller the training error will be;
    \item the bigger $p$ is, the smaller the (cross-domain) generalization error (\textit{i.e.}, difference between testing error and training error) will be. 
\end{itemize}
Therefore, the success of our method depends on the choice of $p$ 
as a balance of the above two goals.


\subsection{Engineering Specification \& Extensions}



For simplicity, we detail the \method{} implementation on a ResNet backbone + FC classification network. 
\method{} is applied to the training phase, and operates on the last convolution feature tensor of ResNet. Denote the feature tensor of an input sample as ${\bf{Z}}$ and its gradient tensor of as ${\bf{G}}$. ${\bf{G}}$ is computed by back propagating the classification score with respect to the ground truth category. Both of them are of size $[7 \times 7 \times 512]$. 

\textbf{Spatial-wise RSC:}
In the training phase, global average pooling is applied along the channel dimension to the gradient tensor ${\bf{G}}$ to produce a weighting matrix  $w_{i}$ of size $[7\times 7]$. Using this matrix, we select top $p$ percentage of the $7\times 7=49$ cells, and mute its corresponding features in ${\bf{Z}}$. 
Each of the $49$ cells correspond to a $[1\times 1\times 512]$ feature vector in ${\bf{Z}}$. 
After that, the new feature tensor ${\bf Z_{new}}$ is forwarded to the new network output. Finally, the network is updated through back-propagation. 
We refer this setup as spatial-wise RSC, which is the default RSC for the rest of this paper. 





\textbf{Channel-wise RSC:}  RSC can also be implemented by dropping features of the channels with high-gradients. The rational behind the channel-wise RSC lies in the convolutional nature of DNNs. The feature tensor of size $[7 \times 7 \times 512]$ can be considered a decomposed version of input image, where instead of the RGB colors, there are $512$ different characteristics of the each pixels. The $C$ characteristics of each pixel contains different statistics of training data from that of the spatial feature statistics. 

For channel-wise RSC, global average pooling is applied along the spatial dimension of ${\bf{G}}$, and produce a weighting vector of size $[1\times 512]$.  Using this vector, we select top $p$ percentage of its $512$ cells, and mute its corresponding features in ${\bf{Z}}$. Here, each of the $512$ cells correspond to a $[7\times 7]$ feature matrix in ${\bf{Z}}$. After that, the new feature tensor ${\bf Z_{new}}$ is forwarded to the new network output. Finally, the network is updated through back-propagation. 

\textbf{Batch Percentage:} 
Some dropout methods like curriculum dropout~\cite{morerio2017curriculum} do not apply dropout at the beginning of training, which improves CNNs by learning basic discriminative clues from unchanged feature maps. Inspired by these methods, we randomly apply RSC to some samples in each batch, leaving the other unchanged. This introduces one extra hyperparameter, namely Batch Percentage: the percentage of samples to apply RSC in each batch. We also apply RSC to top percentage of batch samples based on cross-entropy loss. This setup is slight better than randomness.


Detailed ablation study on above extensions will be conducted in the experiment section below. 


\section{Experiments}
\label{sec:exp}
\subsection{Datasets}

We consider the following four data collections as the battleground to evaluate \method{} against previous methods. 





\begin{itemize}
    \item \textbf{PACS} \cite{li2017deeper}: seven classes over four domains (Artpaint, Cartoon, Sketches, and Photo). The experimental protocol is to train a model on three domains and test on the remaining domain. 
    \item \textbf{VLCS} \cite{torralba2011unbiased}: five classes over four domains. The domains are defined by four image origins, \textit{i.e.}, images were taken from the PASCAL VOC 2007, LabelMe, Caltech and Sun datasets.
    \item \textbf{Office-Home} \cite{venkateswara2017deep}: 65 object categories over 4 domains (Art, Clipart, Product, and Real-World). 
    \item \textbf{ImageNet-Sketch} \cite{wang2019learning}: 1000 classes with two domains. The protocol is to train on standard ImageNet~\cite{ILSVRC15} training set and test on ImageNet-Sketch. 
\end{itemize}






\subsection{Ablation Study}


We conducted five ablation studies on possible configurations for RSC on the PACS dataset~\cite{li2017deeper}. All results were produced based on the ResNet18 baseline in \cite{carlucci2019domain} and were averaged over five runs. 

\textbf{(1)} Feature Dropping Strategies (Table~\ref{table:Ablation1}). We compared the two attention mechanisms to select the most discriminative spatial features. The ``Top-Activiation"~\cite{park2016analysis} selects the features with highest norms, whereas the ``Top-Gradient" (default in RSC) selects the features with high gradients. The comparison shows that ``Top-Gradient" is better than ``Top-Activation", 
while both are better than the random strategy.  Without specific note, we will use ``Top-Gradient" as default in the following ablation study.

\textbf{(2)} Feature Dropping Percentage (choice of $p$) (Table~\ref{table:Ablation2}): We ran RSC at different dropping percentages to mute spatial feature maps. The highest average accuracy was reached at $p=33.3\%$. 
While the best choice of $p$ is data-specific, our results align well with the theoretical discussion: the optimal $p$ should be neither too large nor too small. 

\begin{table}[!htb]
\footnotesize
\centering 
\fontsize{7}{8}\selectfont
\begin{tabular}{c| c| c| c| c| c|| c } 
\hline 
Feature Drop Strategies & backbone & artpaint & cartoon & sketch & photo & Avg $\uparrow$ \\ [0.5ex] 
\hline\hline 
 Baseline~\cite{carlucci2019domain}& ResNet18  & 78.96 & 73.93  & 70.59 & \textbf{96.28} & 79.94 \\ %
 Random & ResNet18  & 79.32 & 75.27 & 74.06 & 95.54 & 81.05\\
 Top-Activation& ResNet18  & 80.31 & 76.05 & 76.13 & \textbf{95.72}&  82.03\\ %
 Top-Gradient& ResNet18  & \textbf{81.23} & \textbf{77.23} & \textbf{77.56} & 95.61& \textbf{82.91} \\ %
\hline 
\end{tabular}
\vspace{0.5em}
\caption{Ablation study of Spatial-wise RSC on Feature Dropping Strategies. Feature Dropping Percentage $50.0\%$ and Batch Percentage $50.0\%$.} 
\vspace{-.5em}
\label{table:Ablation1} 
\end{table}

\begin{table}[!htb]
\footnotesize
\centering 
\fontsize{7}{8}\selectfont
\begin{tabular}{c| c| c| c| c| c|| c } 
\hline 
Feature Dropping Percentage & backbone & artpaint & cartoon & sketch & photo & Avg$\uparrow$  \\ [0.5ex] 
\hline\hline 
 66.7\%& ResNet18  & 80.11 & 76.35 & 76.24 & 95.16 & 81.97 \\
 50.0\%& ResNet18  & 81.23 & 77.23 & 77.56 & 95.61& 82.91 \\ %
 33.3\%& ResNet18  & \textbf{82.87} & \textbf{78.23} & \textbf{78.89} & 95.82& \textbf{83.95} \\
 25.0\%& ResNet18  & 81.63& 78.06 & 78.12&  96.06& 83.46 \\ %
 20.0\%& ResNet18  & 81.22& 77.43 &77.83 & 96.25 & 83.18 \\ %
 13.7\%& ResNet18  & 80.71& 77.18 & 77.12& \textbf{96.36} & 82.84 \\ %
\hline 
\end{tabular}
\vspace{0.5em}
\caption{Ablation study of Spatial-wise RSC on Feature Dropping Percentage. We used ``Top-Gradient" and fixed the Batch Percentage ($50.0\%$) here.} 
\vspace{-1em}
\label{table:Ablation2} 
\end{table}

\begin{table}[!htb]
\footnotesize
\centering 
\fontsize{7}{8}\selectfont
\begin{tabular}{c| c| c| c| c| c|| c } 
\hline 
Batch Percentage & backbone & artpaint & cartoon & sketch & photo & Avg$\uparrow$  \\ [0.5ex] 
\hline\hline 
 50.0\%& ResNet18  & \textbf{82.87} & 78.23 & 78.89 & 95.82& 83.95 \\ %
 33.3\%& ResNet18  & 82.32 & \textbf{78.75} & \textbf{79.56}& 96.05 & \textbf{84.17} \\ %
 25.0\%& ResNet18  & 81.85 & 78.32 & 78.75&  \textbf{96.21}& 83.78 \\ %
\hline 
\end{tabular}
\vspace{0.5em}
\caption{Ablation study of Spatial-wise RSC on Batch Percentage. We used ``Top-Gradient" and fixed Feature Dropping Percentage ($33.3\%$). } 
\vspace{-1em}
\label{table:Ablation3} 
\end{table}

\begin{table}[!htb]
\footnotesize
\centering 
\fontsize{7}{8}\selectfont
\begin{tabular}{c| c| c| c| c| c|| c } 
\hline 
Method & backbone & artpaint & cartoon & sketch & photo & Avg$\uparrow$  \\ [0.5ex] 
\hline\hline 
 Spatial& ResNet18  & 82.32 & 78.75 & 79.56 & \textbf{96.05} & 84.17 \\ %
 Spatial+Channel& ResNet18  & \textbf{83.43} & \textbf{80.31}  & \textbf{80.85} & 95.99 & \textbf{85.15} \\ %
\hline 
\end{tabular}
\vspace{0.5em}
\caption{Ablation study of Spatial-wise RSC verse Spatial+Channel RSC. We used the best strategy and parameter by Table~\ref{table:Ablation3}:``Top-Gradient", Feature Dropping Percentage($33.3\%$) and Batch Percentage($33.3\%$).} 
\vspace{-1em}
\label{table:Ablation4} 
\end{table}

\begin{table}[!htb]
\footnotesize
\centering 
\fontsize{7}{8}\selectfont
\begin{tabular}{c| c| c| c| c| c|| c } 
\hline 
Method & backbone & artpaint & cartoon & sketch & photo & Avg$\uparrow$  \\ [0.5ex] 
\hline\hline 
 Baseline~\cite{carlucci2019domain}& ResNet18  & 78.96 & 73.93  & 70.59 & \textbf{96.28} & 79.94 \\ %
 Cutout\cite{devries2017improved}& ResNet18  & 79.63 & 75.35  & 71.56 & 95.87 & 80.60 \\ %
 DropBlock\cite{ghiasi2018dropblock}& ResNet18  & 80.25 & 77.54  & 76.42 & 95.64 & 82.46 \\ %
 AdversarialDropout\cite{park2018adversarial}& ResNet18  & 82.35 &  78.23 & 75.86 & 96.12 & 83.07 \\ %
 Random(S+C)& ResNet18  & 79.55 & 75.56 & 74.39 & 95.36 &  81.22\\ %
 Top-Activation(S+C)& ResNet18  & 81.03 & 77.86  & 76.65 & 96.11 & 82.91 \\ %
 \textbf{RSC}: Top-Gradient(S+C)& ResNet18  & \textbf{83.43} & \textbf{80.31}  & \textbf{80.85} &95.99  & \textbf{85.15} \\ %
\hline 
\end{tabular}
\vspace{0.5em}
\caption{Ablation study of Dropout methods. ``S" and ``C" represent spatial-wise and channel-wise respectively. For fair comparison, results of above methods are report at their best setting and hyperparameters. RSC used the hyperparameters selected in above ablation studies:``Top-Gradient", Feature Dropping Percentage ($33.3\%$) and Batch Percentage ($33.3\%$). } 
\vspace{-1em}
\label{table:Ablation5} 
\end{table}

\textbf{(3)} Batch Percentage (Table~\ref{table:Ablation3}): RSC has the option to be only randomly applied to a subset of samples in each batch.
Table~\ref{table:Ablation3} shows that the performance is relatively constant. Nevertheless we still choose $33.3\%$ as the best option on the PACS dataset. 

\textbf{(4)} Spatial-wise plus Channel-wise RSC (Table~\ref{table:Ablation4}): 
In ``Spatial+Channel", both spatial-wise and channel-wise RSC were applied on a sample at $50\%$ probability, respectively. (Better options of these probabilities could be explored.) Its improvement over Spatial-wise RSC indicates that it further activated features beneficial to target domains.

\begin{table}[!htb]
\footnotesize
\centering 
\fontsize{7}{8}\selectfont
\begin{tabular}{c| c| c| c| c| c|| c } 
\hline 
PACS & backbone & artpaint & cartoon & sketch & photo & Avg $\uparrow$ \\ [0.5ex] 
\hline\hline 
Baseline\cite{carlucci2019domain} & AlexNet & 66.68 & 69.41 & 60.02 & 89.98 & 71.52 \\ 
Hex\cite{wang2018learning} & AlexNet & 66.80 & 69.70 & 56.20 & 87.90 & 70.20 \\ %
PAR\cite{wang2019learning} & AlexNet & 66.30 & 66.30 & 64.10 & 89.60 & 72.08\\
MetaReg\cite{balaji2018metareg} & AlexNet & 69.82 & 70.35 & 59.26 & \textbf{91.07} & 72.62 \\ %
Epi-FCR\cite{li2019episodic} & AlexNet& 64.70 & 72.30 & 65.00 & 86.10 & 72.00  \\
JiGen\cite{carlucci2019domain} & AlexNet & 67.63 & 71.71 & 65.18 & 89.00 & 73.38 \\ %
MASF\cite{dou2019domain} & AlexNet & 70.35 & 72.46 & \textbf{67.33} & 90.68 & 75.21 \\ %
RSC(ours) & AlexNet & \textbf{71.62} & \textbf{75.11} & 66.62 & 90.88 & \textbf{76.05} \\ %
\hline 
Baseline\cite{carlucci2019domain} & ResNet18 & 78.96 & 73.93 & 70.59 & \textbf{96.28} & 79.94\\ %
MASF\cite{dou2019domain} & ResNet18 & 80.29 & 77.17 & 71.69& 94.99 & 81.03 \\ %
Epi-FCR\cite{li2019episodic} & ResNet18& 82.10 & 77.00 & 73.00 & 93.90 & 81.50  \\
JiGen\cite{carlucci2019domain} & ResNet18 & 79.42 & 75.25 & 71.35 & 96.03 & 80.51 \\ %
MetaReg\cite{balaji2018metareg} & ResNet18 & \textbf{83.70} & 77.20 & 70.30 & 95.50 & 81.70 \\ %
RSC(ours) & ResNet18 & 83.43 & \textbf{80.31}  & \textbf{80.85} & 95.99 & \textbf{85.15} \\ %
\hline 
Baseline\cite{carlucci2019domain} & ResNet50 & 86.20 & 78.70 & 70.63 & 97.66 & 83.29\\ %
MASF\cite{dou2019domain} & ResNet50 & 82.89 & 80.49 & 72.29& 95.01 & 82.67 \\ %
MetaReg\cite{balaji2018metareg} & ResNet50 & 87.20 & 79.20 & 70.30 & 97.60 & 83.60 \\ %
RSC(ours) & ResNet50 & \textbf{87.89} & \textbf{82.16} & \textbf{83.35} & \textbf{97.92}& \textbf{87.83}  \\ %
\hline 
\end{tabular}
\vspace{0.5em}
\caption{DG results on PACS\cite{li2017deeper} (Best in bold).} 
\vspace{-2em}
\label{table:PACS} 
\end{table}

\begin{table}[!htb]
\footnotesize
\centering 
\fontsize{7}{8}\selectfont
\begin{tabular}{c| c| c| c| c| c|| c } 
\hline 
VLCS & backbone & Caltech & Labelme & Pascal & Sun & Avg $\uparrow$ \\ [0.5ex] 
\hline\hline 
Baseline\cite{carlucci2019domain} & AlexNet & 96.25 & 59.72 & 70.58 & 64.51 & 72.76\\ 
Epi-FCR\cite{li2019episodic} & AlexNet& 94.10 & 64.30 & 67.10 & 65.90 & 72.90  \\
JiGen\cite{carlucci2019domain} & AlexNet & 96.93 & 60.90 & 70.62 & 64.30 & 73.19 \\ %
MASF\cite{dou2019domain} & AlexNet & 94.78 & \textbf{64.90} & 69.14 & 67.64 & 74.11 \\ %
RSC(ours) & AlexNet & \textbf{97.61} & 61.86 & \textbf{73.93} & \textbf{68.32} & \textbf{75.43} \\ %
\hline 
\end{tabular}
\vspace{0.5em}
\caption{DG results on VLCS~\cite{torralba2011unbiased} (Best in bold).} 
\vspace{-1em}
\label{table:VLCS} 
\end{table}

\textbf{(5)} Comparison with different dropout methods (Table~\ref{table:Ablation5}): 
Dropout has inspired a number of regularization methods for CNNs. The main differences between those methods lie in applying stochastic or non-stochastic dropout mechanism at input data, convolutional or fully connected layers. Results shows that our gradient-based RSC is better. We believe that gradient is an efficient and straightforward way to encode the sensitivity of output prediction. To the best of our knowledge, we compare with the most related works and illustrate the impact of gradients. 
(a) Cutout~\cite{devries2017improved}. Cutout conducts random dropout on input images, which shows limited improvement over the baseline. 
(b) DropBlock~\cite{ghiasi2018dropblock}. DropBlock tends to dropout discriminative activated parts spatially. It is better than random dropout but inferior to non-stochastic dropout methods in Table~\ref{table:Ablation5} such as AdversarialDropout, Top-Activation and our RSC.
(c) AdversarialDropout~\cite{park2018adversarial,lee2019drop}. AdversarialDropout is based on divergence maximization, while RSC is based on top gradients in generating dropout masks. Results show evidence that the RSC is more effective than AdversarialDropout. 
(d) Random and Top-Activation dropout strategies at their best hyperparameter settings.

\begin{table}[!htb]
\footnotesize
\centering 
\fontsize{7}{8}\selectfont
\begin{tabular}{c| c| c| c| c| c|| c } 
\hline 
Office-Home & backbone & Art & Clipart & Product & Real & Avg $\uparrow$ \\ [0.5ex] 
\hline\hline 
Baseline\cite{carlucci2019domain} & ResNet18 & 52.15 & 45.86 & 70.86 &73.15 & 60.51\\ 
JiGen\cite{carlucci2019domain} & ResNet18 & 53.04 & 47.51 & 71.47 &72.79 & 61.20 \\ %
RSC(ours) & ResNet18 & \textbf{58.42} & \textbf{47.90} & \textbf{71.63} & \textbf{74.54}& \textbf{63.12} \\ %
\hline 
\end{tabular}
\vspace{0.5em}
\caption{DG results on Office-Home~\cite{venkateswara2017deep} (Best in bold).} 
\vspace{-1em}
\label{table:Office} 
\end{table}

\begin{table}[!htb]
\footnotesize
\centering 
\fontsize{7}{8}\selectfont
\begin{tabular}{c| c| c| c } 
\hline 
ImageNet-Sketch & backbone & Top-1 Acc $\uparrow$ & Top-5 Acc $\uparrow$\\ [0.5ex] 
\hline\hline 
Baseline\cite{wang2018learning} & AlexNet & 12.04  & 24.80\\ 
Hex\cite{wang2018learning} & AlexNet & 14.69  & 28.98\\ 
PAR \cite{wang2019learning} & AlexNet & 15.01  & 29.57 \\ 
RSC(ours) & AlexNet & \textbf{16.12}  & \textbf{30.78}\\
\hline 
\end{tabular}
\vspace{0.5em}
\caption{DG results on ImageNet-Sketch~\cite{wang2019learning}. } 
\vspace{-1em}
\label{table:sketch} 
\end{table}

\subsection{Cross-Domain Evaluation}


Through the following experiments, we used ``Top-Gradient" as feature dropping strategy, $33.3\%$ as Feature Dropping Percentages, $33.3\%$ as Batch Percentage, and Spatial+Channel RSC. All results were averaged over five runs. 
In our RSC implementation, we used the SGD solver, $30$ epochs, and batch size $128$. The learning rate starts with $0.004$ for ResNet and $0.001$ for AlexNet, learning rate decayed by 0.1 after 24 epochs. 
For PACS experiment, we used the same data augmentation protocol of randomly cropping the images to retain between $80\%$ to $100\%$, randomly applied horizontal flipping and randomly ($10\%$ probability) convert the RGB image to greyscale, following \cite{carlucci2019domain}. 

In Table.~\ref{table:PACS},\ref{table:VLCS},\ref{table:Office}, we compare RSC with the latest domain generalization work, such as Hex \cite{wang2018learning}, PAR \cite{wang2019learning}, JiGen \cite{carlucci2019domain} and MetaReg \cite{balaji2018metareg}. All these work only report results on different small networks and datasets. 
For fair comparison, we compared RSC to their reported performances with their most common choices of DNNs (\textit{i.e.}, AlexNet, ResNet18, and ResNet50) and datasets. 
\method{} consistently outperforms other competing methods. 


The empirical performance gain of \method{} can be better appreciated if we have a closer look at the PACS experiment in Table.~\ref{table:PACS}. 
The improvement of RSC from the latest baselines~\cite{carlucci2019domain} are significant and consistent: $4.5$ on AlexNet, $5.2$ on ResNet18, and $4.5$ on ResNet50.
It is noticeable that, with both ResNet18 and ResNet50, 
\method{} boosts the performance significantly for sketch domain, 
which is the only colorless domain. 
The model may have to understand the semantics of the object to 
perform well on the sketch domain. 
On the other hand, \method{} performs only marginally better than competing methods
in photo domain, 
which is probably because that photo domain is the simplest one
and every method has already achieved high accuracy on it.




\section{Discussion}
\label{sec:diss}
\begin{table}[!htb]
\footnotesize
\centering 
\fontsize{7}{8}\selectfont
\begin{tabular}{c| c| c| c | c} 
\hline 
ImageNet & backbone & Top-1 Acc $\uparrow$ & Top-5 Acc $\uparrow$& $\#$Param. $\downarrow$  \\ [0.5ex] 
\hline\hline 
Baseline & ResNet50 & 76.13 & 92.86 & 25.6M \\
RSC(ours) & ResNet50 & 77.18 & 93.53 & 25.6M   \\ %
\hline
Baseline & ResNet101 & 77.37 & 93.55 & 44.5M  \\
RSC(ours) & ResNet101 & 78.23 & 94.16 & 44.5M  \\ %
\hline
Baseline & ResNet152 & 78.31 & 94.05 &60.2M   \\
RSC(ours) & ResNet152 & 78.89 & 94.43 &60.2M   \\ %
\hline 
\end{tabular}
\caption{Generalization results on ImageNet. Baseline was produced with official Pytorch implementation and their ImageNet models.}
\label{table:ImageNet} 
\end{table}

\textbf{Standard ImageNet Benchmark:}  
With the impressive performance observed in the cross-domain evaluation, 
we further explore to evaluate the benefit of RSC with other benchmark data and higher network capacity.
 
 
 

We conducted image classification experiments on the Imagenet database\cite{ILSVRC15}. 
We chose three backbones with the same architectural design while with clear hierarchies in model capacities: ResNet50, ResNet101, and ResNet152. 
All models were finetuned for 80 epochs with learning rate decayed by 0.1 every 20 epochs. The initial learning rate for ResNet was 0.01. All models follow extra the same training prototype in default Pytorch ImageNet implementation\footnote{https://github.com/pytorch/examples}, using original batch size of 256, standard data augmentation and $224 \times 224$ as input size.
 
The results in Table~\ref{table:ImageNet} shows that RSC exhibits the ability reduce the performance gap between networks of same family but different sizes (\textit{i.e.}, ResNet50 with RSC approaches the results of baseline ResNet101, and ResNet101 with RSC approaches the results of baseline ResNet151). 
The practical implication is that, RSC could induce faster performance saturation than increasing model sizes. Therefore one could scale down the size of networks to be deployed at comparable performance. 
 

\section{Conclusion}
\label{sec:con}

We introduced 
a simple training heuristic method that can be 
directly applied to almost any CNN architecture 
with no extra model architecture, 
and almost no increment of computing efforts.
We name our method Representation Self-challenging (RSC). 
RSC iteratively forces a CNN to activate features that are less dominant in the training domain, but still correlated with labels. Theoretical and empirical analysis of RSC validate that it is a fundamental and effective way of expanding feature distribution of the training domain. RSC produced the state-of-the-art improvement over baseline CNNs under the standard DG settings of small networks and small datasets. Moreover, our work went beyond the standard DG settings, to illustrate effectiveness of RSC on more prevalent problem scales, \textit{e.g.}, the ImageNet database and network sizes up-to ResNet152.

\clearpage
%
%

\bibliographystyle{splncs04}
\bibliography{ref}


\newpage
\beginsupplement

\onecolumn
\section*{Appendix}

\section{Assumptions}
\begin{itemize}
\setlength{\itemindent}{.1in}
    \item [\textbf{A1}:] $\Theta$ is finite; $l(\cdot,\cdot)$ is zero-one loss for binary classification. 
\end{itemize}
%

The assumption leads to classical discussions on the \textit{i.i.d} setting in multiple textbooks (\textit{e.g.}, \cite{mitchell1997machine}). 
However, modern machine learning concerns more than the \textit{i.i.d} setting, therefore, we need to quantify the variations between train and test distributions. Analysis of domain adaptation is discussed \cite{ben2010theory}, but still relies on the explicit knowledge of the target distribution to quantify the bound with an alignment of the distributions. The following discussion is devoted to the scenario when we do not have the target distribution to align.  

Since we are interested in the $\theta^\star$ instead of the $\theta^\star(\D)$,
we first assume $\Theta$ is large enough and we can find a global optimum hypothesis that is applicable to any distribution, or in formal words:
\begin{itemize}
\setlength{\itemindent}{.1in}
    \item [\textbf{A2}:] $L(\theta^\star;\D) = L(\theta^\star(\D);\D)$ for any $\D$.  
\end{itemize}
This assumption can be met when the conditional distribution $\mathbb{P}(\Y(\D)|\Z(\D))$ is the same for any $\D$. 

\begin{itemize}
\setlength{\itemindent}{.1in}
    \item [\textit{e.g.},] The true concept of ``cat'' is the same for any collection of images. 
\end{itemize}


The challenge of cross-domain evaluation comes in when there exists multiple optimal hypothesis that are equivalently good for one distribution, but not every optimal hypothesis can be applied to other distributions. 

\begin{itemize}
\setlength{\itemindent}{.1in}
    \item [\textit{e.g.},] For the distribution of picture book, ``cats have chubby faces'' can predict the true concept of ``cat''. 
    A model only needs to learn one of these signals to reduce training error, although the other signal also exists in the data. 
\end{itemize}

The follow-up discussion aims to show that \method{} can force the model to learn multiple signals, so that it helps in cross-domain generalization. 



Further, Assumption \textbf{A2} can be interpreted as there is at least some features $\mathbf{z}$ that appear in every distributions we consider. 
We use $i$ to index this set of features. 
Assumption \textbf{A2} also suggests that $\z_i$ is \textit{i.i.d.} (otherwise there will not exist $\theta^\star$) across all the distributions of interest 
(but $\z$ is not \textit{i.i.d.} because $\z_{-i}$, where $-i$ denotes the indices other than $i$, can be sampled from arbitrary distributions).

\begin{itemize}
\setlength{\itemindent}{.1in}
    \item [\textit{e.g.},] $\mathbf{z}$ is the image; $\z_i$ is the ingredients of the true concept of a ``cat'', such as ears, paws, and furs; $\z_{-i}$ is other features such as ``sitting by the window''. 
\end{itemize}

We use $\so$ to specify the distribution that has values on the $i$\textsuperscript{th}, 
but $0$s elsewhere. 
We introduce the next assumption:
\begin{itemize}
\setlength{\itemindent}{.1in}
    \item [\textbf{A3}:] Samples of any distribution of interest (denoted as $\mathcal{A}$) are perturbed version of samples from $\so$ by sampling arbitrary features for $\z_{-i}$: $\mathbb{E}_\mathcal{A}[\mathbb{E}_\s[\z]] = \mathbb{E}_\so[\z]$
\end{itemize}

Notice that this does not contradict with our cross-domain set-up: while Assumption \textbf{A3} implies that data from any distribution of interest is \textit{i.i.d} (otherwise the operation $\mathbb{E}_\mathcal{A}[]$ is not valid), 
the cross-domain difficulty is raised when 
only different subsets of $\mathcal{A}$ are used for train and test. 
For example, 
considering $\mathcal{A}$ to be a uniform distribution of $[0, 1]$, while the train set is uniformly sampled from $[0, 0.5]$ and the test set is uniformly sampled from $(0.5, 1]$. 

\section{Proof of Theoretical Results}
\subsection{Corollary 1}
\begin{proof}
We first study the convergence part, where we consider a fixed hypothesis.
We first expand 
\begin{align*}
    & |L(\etr(\s);\s) - L(\otr(\s); \D)| \\
    & = |L(\etr(\s);\s) - L(\etr(\s);\D) + L(\etr(\s);\D) - L(\otr(\s);\D)| \\
    & \leq |L(\etr(\s);\s) - L(\otr(\s);\D)| + |L(\otr(\s);\D) - L(\otr(\s);\D)|
\end{align*}

We first consider the term $|L(\otr(\s);\s) - L(\otr(\s);\D)|$, 
where we can expand
\begin{align*}
    |L(\otr(\s);\s) - L(\otr(\s);\D)| \leq 2|L(\otr(\s);\s) - L(\otr(\s);\so)|
\end{align*}
because of Assumption \textbf{A4}. 

Also, because of Assumption \textbf{A4}, if samples in $\s$ are perturbed versions of samples in $\so$, then samples in $\so$ can also be seen as perturbed versions of samples in $\s$, thus, Condition~\ref{eq:perturbation} can be directly re-written into:
\begin{align*}
    |L(\otr(\s);\s) - L(\otr(\s);\so)| \leq \xi(p),
\end{align*}
which directly leads us to the fact that $|L(\otr(\s);\s) - L(\otr(\s);\D)|$ has the expectation 0 (\textbf{A4}) and bounded by $[0, \xi(p)]$. 

For $|L(\etr(\s);\s) - L(\otr(\s);\s)|$, 
the strategy is relatively standard. 
We first consider the convergence of a fixed hypothesis $\tr$, then over $n$ \textit{i.i.d} samples, 
the empirical risk ($\widehat{L}(\tr)$) will be bounded within $[0, 1]$ with the expectation $L(\tr)$. 

Before we consider the uniform convergence step, 
we first put the two terms together and apply the Hoeffding's inequality. 
When the random variable is with expectation $L(\tr)$ and bound $[0, 1+2\xi(p)]$,
we have:
\begin{align*}
    \mathbb{P}(|\widehat{L}(\tr;\s)-L(\tr;\D)| \geq \epsilon) \leq 2\exp(-\dfrac{2n\epsilon^2}{(2\xi(p)+1)^2})
\end{align*}
Now, we consider the uniform convergence case, where we have:
\begin{align*}
    \mathbb{P}(\sup_{\tr\in\Tr}|\widehat{L}(\tr;\s)-L(\tr;\D)| \geq \epsilon) \leq 2|\Tr|\exp(-\dfrac{2n\epsilon^2}{(2\xi(p)+1)^2})
\end{align*}
Rearranging these terms following standard tricks will lead to the conclusion. 

\end{proof}

\subsection{Corollary 2}
\begin{proof}
Since we only concern with iteration $t$, we drop the subscript of $\z_t$ and $\zp_t$. 
We first introduce another shorthand notation
\begin{align*}
    h(\etr(t+1), \z) := \sum_{\langle\z_t,\y\rangle} l(f(\z;\etr);\y)
\end{align*}

We expand 
\begin{align*}
     \Gamma(\etr(t+1)) =& |h(\etr(t+1), \z) - h(\etr(t+1), \zp)| \\
     = & |h(\etr(t+1), \z) - h(\etr(t), \zp) + h(\etr(t), \zp) - h(\etr(t+1), \zp)| \\
     = & |h(\etr(t+1), \z) - h(\etr(t), \z) + h(\etr(t), \z) - h(\etr(t), \zp) \\
     & + h(\etr(t), \zp) - h(\etr(t+1), \zp)| \\
     = & |h(\etr(t+1), \z) - h(\etr(t), \z) + h(\etr(t), \zp) - h(\etr(t+1), \zp) + \Gamma(\etr(t))|
\end{align*}

Recall that, by the definition of \method{}, we have:
\begin{align*}
    \etr(t+1) = \etr(t) - \dfrac{\partial h(\etr(t), \zp)}{\partial \etr(t)}\eta = \etr(t) - \gp\eta 
\end{align*}

We apply Taylor expansion over $h(\etr(t+1), \cdot)$ with respect to $\etr(t)$ and have:
\begin{align*}
    h(\etr(t+1), \cdot) = & h(\etr(t), \cdot) + \dfrac{\partial h(\etr(t), \cdot)}{\partial\etr(t)}(\etr(t+1)-\etr(t)) \\
    & + \dfrac{1}{2}\dfrac{\partial^2 h(\etr(t), \cdot)}{\partial^2\etr(t)}||\etr(t+1)-\etr(t)||_2^2 + \sigma \\
    = & h(\etr(t), \cdot) - \dfrac{\partial h(\etr(t), \cdot)}{\partial\etr(t)}\gp\eta
    + \dfrac{1}{2}\dfrac{\partial^2 h(\etr(t), \cdot)}{\partial^2\etr(t)}||\gp\eta ||_2^2 + \sigma,
\end{align*}
where $\sigma$ denotes the higher order terms. 

Assumption \textbf{A6} conveniently allows us to drop terms regarding $\eta^2$ or higher orders, so we have:
\begin{align}
    h(\etr(t), \cdot) - h(\etr(t+1), \cdot) = \dfrac{\partial h(\etr(t), \cdot)}{\partial\etr(t)}\gp\eta
    \label{eq:compare}
\end{align}
Finally, when $\cdot$ is replaced by $\z$ and $\zp$,  

we have:
\begin{align*}
    h(\etr(t), \zp) - h(\etr(t+1), \zp) = \dfrac{\partial h(\etr(t), \zp)}{\partial\etr(t)}\gp\eta = ||\gp||_2^2\eta 
\end{align*}
and
\begin{align*}
    h(\etr(t), \z) - h(\etr(t+1), \z) =& \dfrac{1}{\gamma_t(p)}\dfrac{\partial h(\etr(t), \zp)}{\partial\etr(t)}\gp\eta = \dfrac{1}{\gamma_t(p)}||\gp||_2^2\eta 
\end{align*}
We write these terms back and get
\begin{align*}
    \Gamma(\etr(t+1)) = \big|(\dfrac{1}{\gamma_t(p)}-1)||\gp||_2^2\eta + \Gamma(\etr(t))\big|
\end{align*}
We can simply drop the absolute value sign because all these terms are greater than zero. 
Finally, we rearrange these terms and prove the conclusion. 

\end{proof}

\end{document}